\documentclass{article}
\usepackage[utf8]{inputenc}

\usepackage{xcolor,soul}
\usepackage{cite}
\usepackage{graphicx}
\usepackage{booktabs}
\usepackage{array}
\usepackage{amsfonts, amsmath, amsthm, amssymb}
\usepackage{graphicx}
\usepackage{changepage}
\usepackage{color,soul}
\usepackage{enumitem}   
\usepackage{framed,multirow}
\usepackage{latexsym}
\usepackage{url}
\usepackage{bm}
\usepackage{relsize}

\usepackage[ruled,vlined]{algorithm2e}

\newcolumntype{L}[1]{>{\raggedright\let\newline\\\arraybackslash\hspace{0pt}}m{#1}}
\newcolumntype{C}[1]{>{\centering\let\newline\\\arraybackslash\hspace{0pt}}m{#1}}
\newcolumntype{R}[1]{>{\raggedleft\let\newline\\\arraybackslash\hspace{0pt}}m{#1}}

\usepackage{pifont}

\usepackage{tikz}

\makeatletter
\newcommand{\thickhline}{%
    \noalign {\ifnum 0=`}\fi \hrule height 1pt
    \futurelet \reserved@a \@xhline
}
\newcolumntype{"}{@{\hskip\tabcolsep\vrule width 1pt\hskip\tabcolsep}}
\makeatother

\usepackage{geometry}
\title{Anatomically Constrained Tractography of the Fetal Brain}

\author{Camilo Calixto, Camilo Jaimes, Matheus D. Soldatelli, Simon K. Warfield, \\ Ali Gholipour, and Davood Karimi \\
\\
Computational Radiology Laboratory (CRL), Department of Radiology,\\ Boston Children's Hospital, and Harvard Medical School, USA}

\begin{document}

\maketitle

\begin{abstract}

Diffusion-weighted Magnetic Resonance Imaging (dMRI) is increasingly used to study the fetal brain in utero. An important computation enabled by dMRI is streamline tractography, which has unique applications such as tract-specific analysis of the brain white matter and structural connectivity assessment. However, due to the low fetal dMRI data quality and the challenging nature of tractography, existing methods tend to produce highly inaccurate results. They generate many false streamlines while failing to reconstruct streamlines that constitute the major white matter tracts. In this paper, we advocate for anatomically constrained tractography based on an accurate segmentation of the fetal brain tissue directly in the dMRI space. We develop a deep learning method to compute the segmentation automatically. Experiments on independent test data show that this method can accurately segment the fetal brain tissue and drastically improve tractography results. It enables the reconstruction of highly curved tracts such as optic radiations. Importantly, our method infers the tissue segmentation and streamline propagation direction from a diffusion tensor fit to the dMRI data, making it applicable to routine fetal dMRI scans. The proposed method can lead to significant improvements in the accuracy and reproducibility of quantitative assessment of the fetal brain with dMRI.

\end{abstract}

\section{Introduction}

\subsection{Background and motivation}

The human brain undergoes dramatic microstructural and macrostructural developments in utero \cite{silbereis2016cellular, bayer2005human}. It can be argued that the fetal period is the most dynamic and critical stage in brain development \cite{collin2013ontogeny, dubois2014early}. Processes such as neurogenesis, synapse formation, neural migration, and axonal growth all begin before birth. These processes form the brain microstructure and lay the foundations for the formation and development of the structural brain connectome \cite{takahashi2012emerging, song2017human}, which will continue its rapid development in the first few months and years after birth. Various diseases and insults can interrupt normal fetal brain development and result in lifelong neurodevelopmental and psychiatric disorders \cite{donofrio2011impact, lynch2009epidemiology, linnet2003maternal}. Therefore, quantitative assessment of normal and abnormal fetal brain development is very useful. Not only can it enhance our understanding of the development of the brain's cognitive capabilities, but it can also facilitate the diagnosis, management, and treatment of neurological disorders at their earliest stage \cite{song2018accurate, huang2009anatomical, kasprian2013assessing, xu2014radial, millischer2022feasibility}.

Diffusion-weighted magnetic resonance imaging (dMRI) has played an increasingly prominent role in studying the development of fetal brain \cite{ouyang2019delineation, huang2006white, huang2009anatomical, jakab2015disrupted}. Although it lacks the spatial resolution of postmortem histological analysis, dMRI is non-invasive, significantly faster and cheaper, enables assessment of the whole brain in utero in 3D, and allows studying large populations. In addition to assessing local tissue microstructure, fetal dMRI can also be used to trace virtual streamlines with tractography techniques. The streamlines provide a visual and quantitative representation of the nerve fibers connecting different areas of the central nervous system. Two important practical applications of tractography include tract-specific analysis of the brain white matter and quantitative structural connectivity assessment.

\begin{itemize}

\item Tractography-generated streamlines can be used to identify and study specific white matter tracts. Accurate delineation of these tracts is needed in clinical studies and medical research. For example, changes in the micro-structural properties of specific tracts are commonly used in studying brain development and disorders. The majority of manual and automatic methods for segmenting these tracts depend on tractography \cite{suarez2012automated, garyfallidis2018recognition, siless2020registration}.

\item Tractography is also the main computation in structural connectivity analysis. Structural connectivity is widely used in neuroscience and medicine to study normal and abnormal brain development \cite{collin2013ontogeny, bullmore2011brain}. The accuracy and reproducibility of quantitative structural connectivity depend greatly on the accuracy of tractography.

\end{itemize}

Although dMRI-based tractography suffers from significant challenges, constant technical advancements have improved its accuracy and reproducibility \cite{sotiropoulos2019building, jones2013white}. Moreover, our understanding of the capabilities and limitations of tractography have improved \cite{jones2010challenges, yang2021diffusion}. As a result, tractography has been increasingly employed in medical applications and neuroscience research \cite{glasser2008dti, essayed2017white, ciccarelli2008diffusion, lo2010diffusion}. However, most recent technical developments and medical applications have focused on post-natal and adult brains. Comparatively, fetal tractography has received far less attention. Fetal brain dMRI suffers from persistent challenges such as low signal-to-noise ratio, imaging artifacts, short scan times, unpredictable motion, and rapid brain development \cite{tymofiyeva2014structural, khan2019fetal, wilson2021development}. These factors have made it challenging to analyze fetal dMRI data and to compute the desired results with the same level of accuracy, detail, and reproducibility as in adult brains. Consequently, compared with post-natal and adult brain imaging, fetal dMRI has remained at a primitive stage \cite{dubois2014early, jakab2017utero, qiu2015diffusion}. In particular, in-utero fetal brain tractography has been much less utilized. 

Most prior works on fetal tractography have focused on one or a few selected tract bundles and have shown low success rates and limited accuracy in reconstructing the full extent of the tracts \cite{song2018accurate, kasprian2008utero, mitter2015vivo, zanin2011white, takahashi2014development, kolasinski2013radial, ouyang2019delineation, huang2006white, huang2009anatomical, jakab2015disrupted, jakab2017utero}. They have mostly applied standard tractography methods with streamline propagation and stopping criteria based on ad-hoc rules such as turning angles or thresholds on fractional anisotropy (FA). This approach can be especially suboptimal for the developing brain, where white matter tracts mature at different rates. The shape and microstructural makeup (hence diffusivity values and FA) of white matter tracts change over time and at different rates for different tracts \cite{vasung2019exploring, kolasinski2013radial, song2015asymmetry, takahashi2012emerging}. To cope with this difficulty, it has been common to empirically adjust FA and angle thresholds in ad-hoc ways to reconstruct different tracts \cite{millischer2022feasibility, mitter2015vivo}. Many studies have resorted to painstaking manual placement of multiple regions of interest (ROIs) to ensure the desired tracts are reconstructed and reduce false positive streamlines \cite{song2015asymmetry, millischer2022feasibility, jaimes2020vivo, kasprian2008utero}. High-quality anatomical images and/or diffusion parameters maps are needed for precise ROI placement \cite{mitter2015vivo, zanin2011white, machado2021tractography}. However, even for postmortem fetal brain tissue, where higher quality dMRI data can be acquired, existing tractography methods tend to produce incomplete and noisy results and often fail to reconstruct the full extent of even the major white matter tracts \cite{takahashi2014development, song2015asymmetry, takahashi2012emerging, kolasinski2013radial, vasung2010development}.

Despite their limitations, prior works have demonstrated that tractography has a unique potential for studying normal and abnormal brain development in utero \cite{huang2009anatomical, wilson2021development, kasprian2008utero, mitter2015vivo, zanin2011white, mitter2015validation, xu2014radial}. Past studies have successfully used tractography of the fetal brain to (1) chart the normal maturation of different white matter tracts and to gain new insights into the timing and speed of tract-specific microstructural changes in utero \cite{zanin2011white, hooker2020third, machado2021spatiotemporal}; (2) to analyze structural connectivity of the fetal brain \cite{limperopoulos2010brain, jaimes2020association}; and (3) to characterize, classify, and understand the clinical heterogeneity of various brain malformations that begin in the fetal period such as agenesis of corpus callosum \cite{jakab2015disrupted, meoded2011prenatal, kasprian2013assessing, millischer2022feasibility}. Hence, there is a critical and urgent need for methods and resources to enable accurate and reproducible tractography of the fetal brain. Accurate tractography can advance the field of fetal brain imaging in multiple ways by facilitating qualitative and quantitative assessment of the development of white matter tracts and of the structural connectome.

\subsection{Contributions of this work}

This paper introduces a new method for fetal brain tractography that leverages tissue segmentation to ensure reconstruction of anatomically valid streamlines. Accurate tissue segmentation enables the implementation of precise streamline seeding and stopping rules that effectively remove spurious streamlines. A deep learning method is developed to compute accurate tissue segmentations. Streamline tracking is performed based on local fiber orientations computed with a diffusion tensor model, which makes the method applicable to single-shell dMRI scans. We tested our method on a set of independent scans and found that it produced significantly more accurate results than a standard tractography technique. Unlike recent studies that rely on unique, high-quality data from the Human Connectome Project \cite{wilson2023spatiotemporal, wilson2021development}, our method can be applied to more typical dMRI scans. We release our software to enable other research groups to use this method as a benchmark for future studies.

\section{Methods}

\subsection{Imaging data acquisition and preprocessing}

In-utero fetal MRI scans for this work were acquired as part of a prospective research study at Boston Children's Hospital in Boston, MA, between 2013 and 2019. The research was approved by the Institutional Review Board (IRB) and adhered to the Health Insurance Portability and Accountability Act (HIPAA). Participants were recruited in advance and provided written consent before each fetal MRI examination. Only pregnancies between 23 and 36 weeks of gestational age (GA) with mothers between 18 and 45 years were included in this work. The exclusion criteria included contraindications to MRI, any form of high-risk pregnancy, fetal brain anomalies, and maternal comorbidities such as diabetes, hypertension, or substance abuse.

Three-Tesla (3T) MRI scanners (Skyra and Prisma, Siemens Medical Solutions) with 16-channel body matrix and spine coils were used to obtain the images. The field of view and number of slices varied based on maternal and fetal dimensions. T2-weighted Half-Fourier Single Shot Turbo Spin Echo (HASTE) fetal brain images were obtained for structural sequences. We acquired multiple images in orthogonal planes with the following acquisition parameters: TR of 1400-2000 ms, TE of 100-120 ms, in-plane resolution of 0.9-1.1 mm, 2 mm slice thickness with no inter-slice space, and 2- or 4-slice interleaved acquisition. For diffusion-weighted images, we obtained 2-8 echo-planar diffusion-weighted images, each along orthogonal planes with respect to the fetal head. Each of these acquisitions included one or two b=0 s/mm\textsuperscript{2} images and 12-24 diffusion-sensitized images at b=500 s/mm\textsuperscript{2}. The acquisition parameters for diffusion-weighted images were TR of 3000-4000 ms, TE of 60 ms, in-plane resolution of 2 mm, and slice thickness of 2-4 mm.

All studies were assessed for the availability of at least two diffusion-weighted scans without excessive motion-induced artifacts, distortion, and signal loss that could hinder successful motion correction and registration between diffusion-weighted and structural T2-weighted scans. We used a previously validated pipeline \cite{khan2019fetal} to process each subject's diffusion and structural MRI data. This pipeline includes the reconstruction of structural and diffusion-weighted volumes with a Kalman filtering-based motion tracking and slice-to-volume registration algorithm \cite{marami2016motion}. It registers the dMRI measurements for each subject into a standard atlas space and produces consistent measurements that consist of scattered q-space data in each voxel. We chose an isotropic voxel size of 1.2 mm for the reconstructed dMRI data. These motion-corrected dMRI volumes were used in the subsequent analysis steps described below.

We excluded subjects that were reconstructed with significant errors or unusually low data quality. This decision was made by an expert via visual inspection of mean diffusivity (MD), color fractional anisotropy maps, and diffusion tensor glyphs to ensure adequate direction of the principal eigenvectors. As a result, 85 fetuses were used to develop and validate the new methods. Data from 74 subjects were used to develop the methods, for hyper-parameter tuning of the deep learning method, and for preliminary evaluations. A total of 11 fetuses were kept separate for independent final validation of the method.

\begin{figure}[!ht]
\centering
\includegraphics[width=150mm]{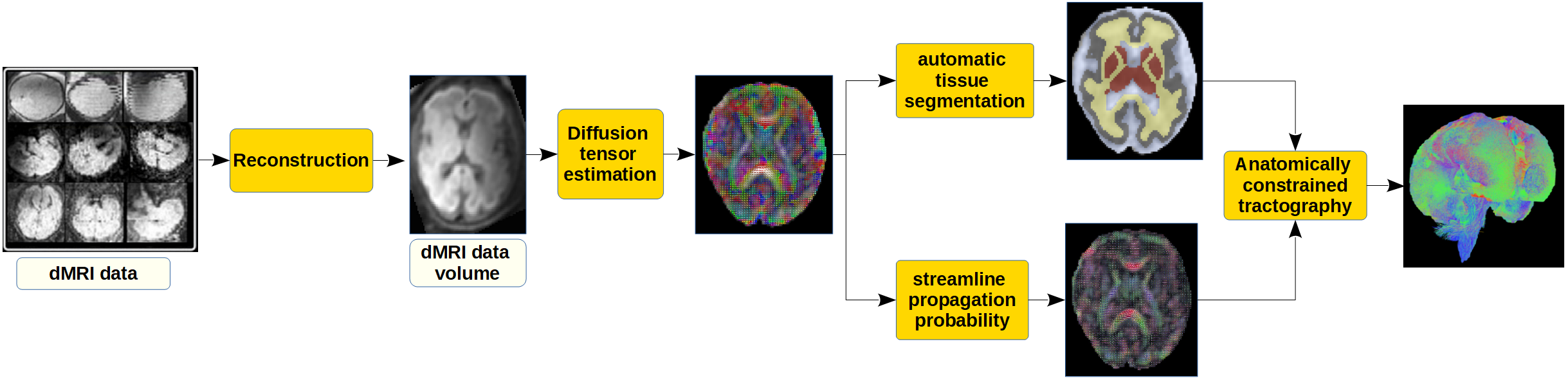}
\caption{A schematic representation of the data processing steps to compute a whole-brain tractogram in this work. Super-resolved reconstruction of dMRI data volumes is performed using the method described in \cite{marami2016motion}.}
\label{fig:pipeline}
\end{figure}

Figure \ref{fig:pipeline} shows a schematic representation of the data processing and computational steps that are applied to compute a whole-brain tractogram in this work. The main steps include local fiber orientation estimation, tissue segmentation, and anatomically constrained tractography, which are described in the following sections.

\subsection{Estimation of local fiber orientation}

Streamline tractography builds upon a voxel-wise estimation of local fiber orientations. It is well known that the accuracy of fiber orientation estimation is one of the main factors influencing the accuracy of tractography \cite{rheault2020common, yeh2021mapping}. For post-natal and adult brains, this need has given rise to advanced methods to compute complex fiber configurations in each voxel, such as crossing fibers and asymmetric fiber orientation distributions \cite{seunarine2014, feng2020asymmetric}. For fetal dMRI, because of the far lower measurement quality, the application of these methods is challenging and rarely attempted. Recently, dense multi-shell data from the Human Connectome Project have been used by a few works to estimate the fiber orientations in the fetal brain using more complex models that the diffusion tensor \cite{pietsch2020multi, deprez2019higher}. However, with standard fetal dMRI scans, the diffusion tensor is the most complex model that can be reliably estimated \cite{dubois2014early, khan2019fetal, jakab2017utero}. Therefore, we used the tensor model to estimate local fiber orientations in this work.

Given the diffusion signal in a voxel, we estimate the diffusion tensor in that voxel using a weighted linear least squares method \cite{koay2006unifying}. We then computed the diffusion orientation distribution function (dODF) from the diffusion tensor ($D$) using the following equation \cite{aganj2010reconstruction, descoteaux2008high}.

\begin{equation}  \label{eq:dODF}
\text{dODF}(u)= \frac{1}{4 \pi | D |^{1/2} \big( u^T D u \big)^{3/2} }
\end{equation}

\noindent
In this equation, $u$ is the unit vector denoting the orientation along which the dODF is computed, and $|D|$ denotes the determinant of $D$.

\begin{figure}[!ht]
\centering
\includegraphics[width=150mm]{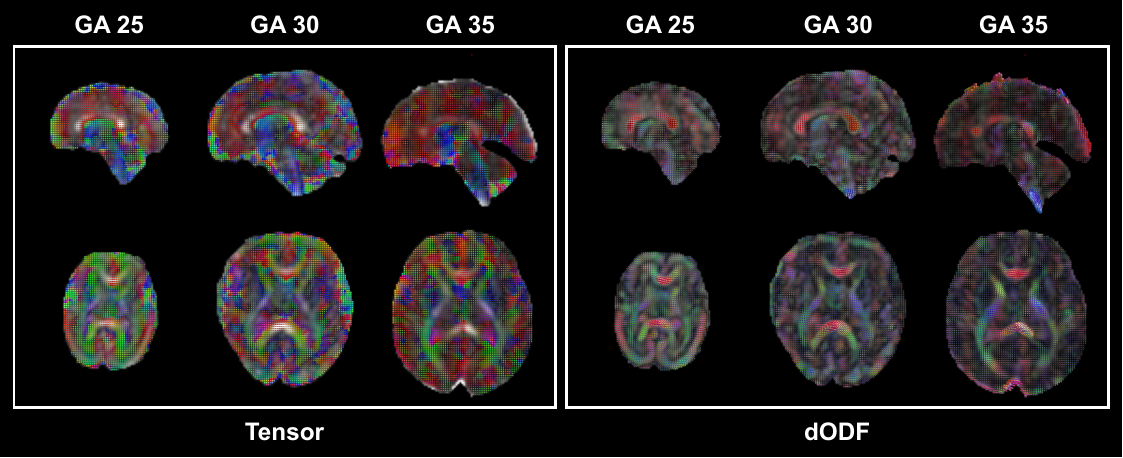}
\caption{Example diffusion tensors and the corresponding streamline propagation probability maps ($p(u)$) for three different fetuses at three different gestational ages (GA).}
\label{fig:fod_sharpening}
\end{figure}

Low diffusion anisotropy in fetal white matter leads to a high prevalence of spurious streamlines with standard tractography techniques. To alleviate this problem, we introduce an ad-hoc but simple and effective method to sharpen the estimated dODF. Specifically, we compute the streamline propagation probability as $p(u)= \text{dODF}^k(u)$, i.e., by raising the dODF to power $k>1$. This operation increases the relative magnitude of $p$ along the directions where dODF is larger and reduces its relative magnitude along the directions where dODF is smaller, thereby effectively increasing the anisotropy of $p$. Figure \ref{fig:fod_sharpening} shows example diffusion tensors and streamline propagation probability maps computed with this approach. We compute $p(u)$ for unit vectors $u$ on a discrete sphere with 724 uniformly-spaced points.

\subsection{Segmentation of the brain tissue}

Anatomically constrained tractography requires accurate brain tissue segmentation in the dMRI space \cite{smith2012anatomically}. Given the lower spatial resolution and contrast of dMRI, this segmentation is usually obtained via segmenting the anatomical MRI and registering the segmentation to the dMRI space. However, the registration is challenging because the contrast, spatial resolution, and image distortions can differ greatly between the two modalities. Registration errors can significantly impact the accuracy of tractography. Therefore, direct tissue segmentation based on the dMRI data is highly desirable. However, this task is challenging due to the varying contrast of dMRI acquisitions, low SNR, and low spatial resolution. To tackle these challenges, classical machine learning methods such as fuzzy c-means clustering with spatial constraints \cite{wen2013brain}, sparse representation of the dMRI signal in dictionaries \cite{yap2015brain}, and support vector machines \cite{ciritsis2018automated} have been used in prior works to address this task. More recently, deep learning methods have been attempted by several studies and have shown better results \cite{golkov2016q, zhang2021deepseg, zhang2020deepb, zhang2015deep}. However, all these prior works have been reported for post-natal and adult brains.

Due to the inferior image quality, rapid brain development, and paucity of training data, segmentation of the fetal brain tissue is significantly more challenging \cite{makropoulos2018review}. Existing works have focused on the segmentation of anatomical MRI images. Several research groups have performed segmentation of the cortical gray matter on T2 images \cite{dou2020deep, dumast2021segmentation}. Segmentation of other tissue types and structures, such as the white matter, ventricles, and cerebellum, has also been addressed by a few recent studies \cite{fidon2021distributionally, payette2023fetal, karimi2023learning}. However, segmentation within the dMRI space remains unexplored in the literature. To address this technology gap, we developed and validated a deep learning method for fast, accurate, and reproducible segmentation of fetal brain tissue directly in dMRI.

\subsubsection{Generation of training and test labels}

\begin{figure}[!ht]
\centering
\includegraphics[width=150mm]{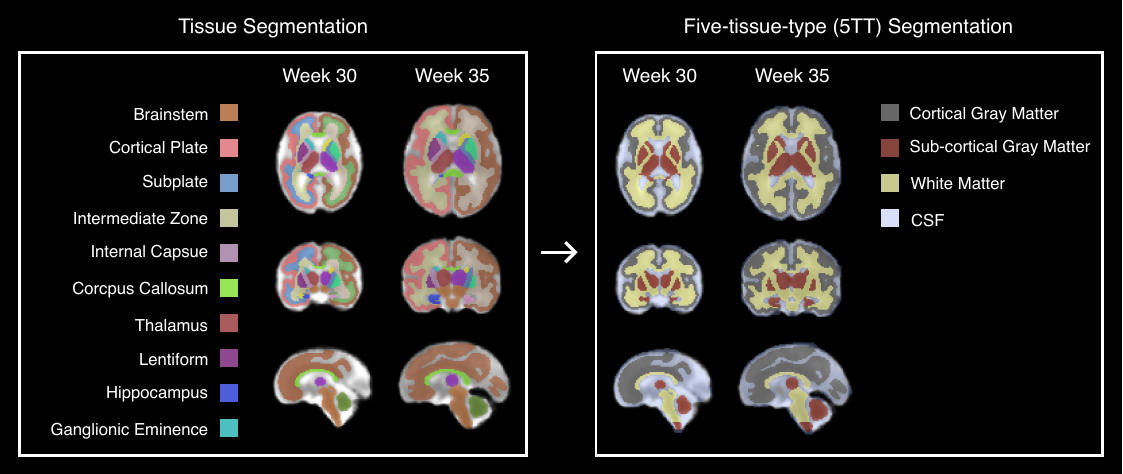}
\caption{Depiction of two diffusion tensor atlases at two different gestational weeks, namely 30 and 35 weeks. The left panel shows manually annotated labels, while the right panel shows the converted labels to the five-tissue-type (5TT) format. The 5TT image is used for anatomically constrained tractography. Please note that the bottom part of the brainstem is converted to a ``gray matter'' to ensure adequate termination of the streamlines in tractography computation.}
\label{fig:labels}
\end{figure}

To create a training dataset, we adopted a multi-atlas segmentation strategy. Our methodology entailed leveraging pre-existing segmentations on high-quality diffusion tensor atlases of fetal brains \cite{calixto2023detailed}. These atlases have been computed at one-week intervals between 23 and 35 gestational weeks and comprise 12 structures/labels for fetuses under 31 gestational weeks and 11 labels for fetuses aged 31 weeks and above, as shown in Figure \ref{fig:labels}. This difference stems from the presence of two transient zones in younger fetuses, namely the subplate and intermediate zones. Note that due to spatial resolution constraints, in our labels, the intermediate zone also contains the ventricular zone and the subventricular zone. Using these atlases, for each fetus, we generated the segmentations by following these steps:

\begin{enumerate}

\item The three closest atlases, in terms of gestational weeks (that is, atlases at weeks $t-1$, $t$, and $t+1$, for a fetus that is closest to age $t$) were registered to the subject's brain. This was performed using a deformable diffusion tensor-based registration technique \cite{zhang2006deformable}.

\item The computed registration transformations were used to align the atlas tissue segmentations to the subject fetal brain using a generic interpolator for labeled images \cite{schaerer2014generic}. The probabilistic STAPLE algorithm \cite{akhondi2013simultaneous} was then applied to fuse these segmentations into a final tissue segmentation prediction for the fetus.

\item Automatic segmentations were reviewed and revised as needed by a board-certified neuroradiologist with fellowship training in pediatric neuroradiology. We excluded automatic segmentations with significant errors, and we ensured the accuracy of the remaining segmentations by having two research fellows with medical training carry out manual refinements where needed.

\item The resulting tissue segmentation map is then converted to the five-tissue-type (5TT) maps that include five labels: white matter, cortical gray matter, sub-cortical gray matter, CSF, and any pathological tissue. Example tissue segmentation maps and 5TT maps are shown in Figure \ref{fig:labels}. The 5TT maps are used for anatomically constrained tractography.

\end{enumerate}

The above four-step procedure was used to generate segmentation labels for the 74 fetuses in the training set. For the 11 fetuses in the independent test set, on the other hand, an expert manually generated the tissue segmentations by following the same approach as that used for the atlases \cite{calixto2023detailed}. This ensured optimal ground truth labels for validation of the new methods.

\subsubsection{Deep learning model development}

An important first decision in developing the automatic tissue segmentation model is the choice of the model input. For this application, we decided to compute the model input from the diffusion tensor, $D$, as $u= v_1(D) \text{FA}(D)$. In this equation, $v_1$ is the eigenvector associated with the largest eigenvalue and $\text{FA}$ is fractional anisotropy. Hence, the input is a three-channel image where each voxel shows the orientation of the highest diffusion, modulated by fractional anisotropy. This representation nicely summarizes the diffusion tensor. The decision to compute the model input from the diffusion tensor has the advantage that it improves the method's generalizability to scans with varying numbers of measurements and different data preprocessing pipelines. Given the fact that the diffusion tensor can be reliably estimated with typical fetal scans, this choice maximizes the method's applicability.

\begin{figure}[!ht]
\centering
\includegraphics[width=153mm]{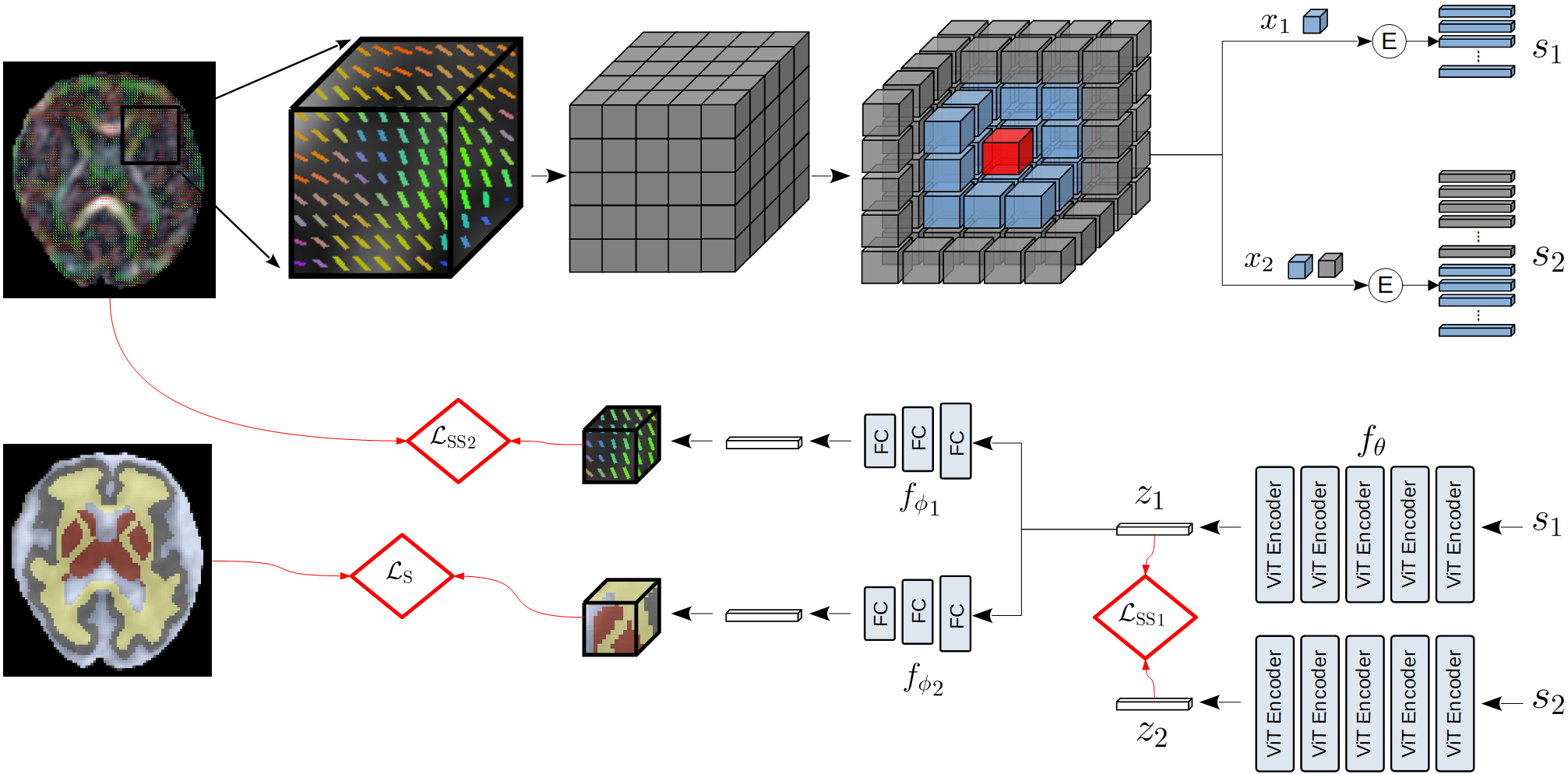}
\caption{Schematic representation of the deep learning model used for automatic fetal brain tissue segmentation.}
\label{fig:network_architecture}
\end{figure}

Figure \ref{fig:network_architecture} shows our proposed segmentation method. It works on image cubes that are partitioned into patches of size $s^3$ voxels. The prediction target is the tissue classification map for the center patch, shown in red in the figure. As input, the network uses the 3-patch and 5-patch neighborhood of the target patch location. Note that the target (center) patch itself is not included in the input context. Therefore, $3^3-1$ and $5^3-1$ patches are used as input to the network. 

We can interpret our approach as providing two separate ``views'' of the input image to the network for performing the desired prediction task. From this viewpoint, our approach is similar to the standard approaches in self-supervised learning methods. However, while the common approach in self-supervised learning is to use data augmentation to generate two views of the same data sample, we use input contexts of different sizes. As described below, our method encourages the network to learn similar representations from both input contexts (one larger than the other) for predicting tissue classification.

The network itself consists of encoder and decoder sections. The encoder includes five standard vision transformer (ViT) encoder modules \cite{dosovitskiy2020image}. The two decoder sections are 3-layer MLPs of fully-connected layers with ReLU activations. We denote the $3^3-1$- and $5^3-1$-patch inputs as $x_1$ and $x_2$, respectively. As shown in Figure \ref{fig:network_architecture}, each input patch is first projected into ${\rm I\!R}^{m}$, followed by fixed sinusoidal positional encodings similar to \cite{vaswani2017attention}. This creates two input sequences $S_1$ and $S_2$ that are passed through the encoder $f_{\theta}$ to compute the representations $z_1$ and $z_2$. The representation $z_1$ is passed to the two decoders: (1) ${f_{\phi}}_1$ computes the input image content (i.e., the FA-weighted direction of the major eigenvector) for the center patch; (2) ${f_{\phi}}_2$ computes the tissue classification for that patch.

\paragraph{Supervised training.} The supervised objective is to achieve accurate tissue classification. We use the cross-entropy loss for this purpose:

\begin{equation} \label{eq:s_loss}
\mathcal{L}_{\text{S}}(x_1, x_2, y | \theta, \phi_2 )= \text{X}_{\text{ent}} (\hat{y}_c,y_c).
\end{equation}

\noindent
where $\hat{y}_c$ and $y_c$ represent the predicted and ground truth tissue classification maps for the center patch. In both Equations \eqref{eq:ss_loss_2} and \eqref{eq:s_loss}, the total loss for each patch is the voxel-wise sum for all voxels in the patch.

\paragraph{Self-supervised training.} The self-supervised objective that we propose encourages: (1) the representations ($z_1$ and $z_2$) computed by the encoder for the two input contexts ($x_1$ and $x_2$) to be similar; and (2) the learned representations are useful for predicting the masked input. We achieve this using the following loss functions:

\begin{equation} \label{eq:ss_loss_1}
\mathcal{L}_{\text{SS}_1} ( x_1, x_2 | \theta )= \Big(1- \cos(z_1^*,z_2^*) \Big) + \alpha \sum_{i} \sum_{j \neq i} C_{i,j}^2
\end{equation}

\begin{equation} \label{eq:ss_loss_2}
\mathcal{L}_{\text{SS}_2} ( x_1, x_2 | \theta, \phi_1)= \| \hat{x}_c -  x_c \|.
\end{equation}

The first term in $\mathcal{L}_{\text{SS}_1}$, based on cosine distance, encourages high similarity between the representations, where $z_1^*$ and $z_2^*$ are $\ell_2$-normalized $z_1$ and $z_2$. The second term is similar to loss functions proposed by canonical correlation analysis-based techniques such as Barlow Twins \cite{zbontar2021barlow} and VICReg \cite{bardes2021vicreg}. In this term, $C$ is the cross-correlation of the representations, computed along the batch dimension:

\begin{equation} \label{eq:x_corr}
C_{i,j} = \frac{\sum_{k=1:K} {z_1^*}^k_i {z_2^*}^k_j}{ \sqrt{ \sum_{k=1:K} \big({z_1^*}^k_i \big)^2} \sqrt{ \sum_{k=1:K}  \big({z_2^*}^k_j\big)^2} },
\end{equation}

\noindent
where $k$ denotes the data sample index within a batch and $K$ is the batch size. This term encourages decorrelation of different elements of the representations computed from the two input contexts. It helps avoid dimensional collapse, which is a common pitfall in self-supervised learning. As in \cite{zbontar2021barlow}, we normalize the representations such that they have a mean of zero across the batch size and a unit norm for each data sample.

The second self-supervised loss, $\mathcal{L}_{\text{SS}_2}$, encourages accurate prediction of the center image patch. In this equation, $\hat{x}_c$ and $x_c$ represent, respectively, the predicted and true image content for the center patch.
 
The combined loss function used to train the model is a weighted sum of the supervised and self-supervised loss terms:

\begin{equation} \label{eq:loss_total}
\mathcal{L}_{\text{total}} ( x_1, x_2, y | \theta, \phi_1, \phi_2 )= \mathcal{L}_{\text{S}} + \lambda_1 \mathcal{L}_{\text{SS}_1} + \lambda_2 \mathcal{L}_{\text{SS}_2}.
\end{equation}

The inclusion of self-supervised training objectives has several advantages. First, based on the results of recent works in computer vision, they are expected to improve the performance on the main task, i.e., tissue classification. Second, self-supervised learning has been shown to be an effective technique for reducing the impact of label noise \cite{zheltonozhskii2022contrast, ghosh2021contrastive}. This is an especially important consideration in this work because the inherently low contrast in fetal dMRI results in inevitable label noise, even in manual annotations. Moreover, self-supervised learning enables us to use additional unlabeled data. In this work, we used 129 preterm newborn subjects scanned between 26 and 36 gestational weeks from the developing Human Connectome Project (dHCP) dataset \cite{bastiani2019}. For training with these unlabeled data, we dropped the supervised loss term $\mathcal{L}_{\text{S}}$ from the combined loss $\mathcal{L}_{\text{total}}$.

We empirically set the hyper-parameters as follows: $\alpha= 0.05$, $\lambda_1= \lambda_2= 0.1$, and patch side length $s=5$ voxels. To train the network using the combined loss, we used the Adam optimization method \cite{kingma2014} with an initial learning rate of $0.001$. We reduced the learning rate by a factor of 0.90 after a training epoch if the validation loss did not decrease. Training with the labeled and unlabeled data was performed in tandem and alternately. Specifically, odd training iterations used a batch of unlabeled data from the dHCP, and even training iterations used a batch of labeled training data from our fetal scans. We used a batch size of 50. The network was implemented in TensorFlow 1.4 and run on an NVIDIA RTX A6000 GPU on a Linux machine with 128 GB of memory and 20 CPU cores.

\subsection{Streamline tracing}

The method developed in this work was based on the anatomically-constrained tractography framework \cite{smith2012anatomically}. For streamline propagation, we use the 2\textsuperscript{nd}-order integration over fibre orientation distributions (iFOD2) algorithm \cite{tournier2010improved} implemented in MRtrix3 \cite{tournier2019mrtrix3}. Streamlines are launched at the boundary of white matter and (cortical and sub-cortical) gray matter. Any streamline that violates the rules of anatomically-constrained tractography is discarded. The main steps of the algorithm are summarized in Algorithm \ref{alg:FetalACT}.

\begin{algorithm} \label{alg:FetalACT}
\caption{The proposed method for anatomically constrained fetal tractography.}\label{alg:fetal_act}
\vspace{1mm}
\textbf{Input:} $\hspace{2mm}$ dMRI data $x$ \& the desired number of streamlines $M$.\\
\textbf{Output:} set of streamlines $\{ y_i \}_{i \in 1, M}$.\\
\vspace{1mm}
Compute the diffusion tensor map $D$ from $x$.\\
Compute the streamline propagation probability $p(u)= \text{dODF}^k(u)$ using Equation \eqref{eq:dODF}.\\
Compute the five-tissue-type image with the deep learning method.\\
Derive a map of the gray matter-white matter interface from the five-tissue-type image.\\
Set minimum and maximum streamline lengths using Equation \eqref{eq:StreamlineLength}.\\
set $y=\{\}$ and $\textit{valid\textunderscore streamline\textunderscore count}= 0$ \\
\While{$\text{valid\textunderscore streamline\textunderscore count}<M$}{
Select a random seed point on the gray matter / white matter boundary. \\
Trace a streamline using iFOD2 until an endpoint is reached. \\
\If { streamline is valid }{
Append the new streamline to $y$. \\
Increment $\textit{valid\textunderscore streamline\textunderscore count} $ by 1. \\
}
}
\vspace{1mm}
\end{algorithm}

For streamline propagation with iFOD2 we used a step size of 0.6mm and angle threshold of $20^{\circ}$. We aimed to reconstruct various projection, association, and commissural fibers while minimizing the number of U-fibers. As the range of valid streamline lengths varies depending on the brain size, we set the maximum streamline length based on the total brain volume (in $mm^3$) to reduce overly long erroneous streamlines. This strategy was helpful in avoiding unrealistic but anatomically valid streamlines (e.g., a streamline that starts at the brainstem, traverses the internal capsule, crosses over to the contralateral hemisphere via the corpus callosum, and then loops back to the brainstem through the contralateral internal capsule). On the other hand, setting a proper minimum length effectively reduced short association fibers, which typically do not contribute to major fiber bundles. The lower and upper thresholds were determined by initially performing global tractography on a few selected subjects and examining the results. Initially, the minimum streamline length was set at 15mm and the maximum at 190mm. After extracting 11 tracts, as listed below, we recorded the shortest streamline length of each tract for every subject. The lowest value among all tracts for every subject was then chosen as the minimum length threshold. Following that, we recorded the 95th percentile of streamline lengths for every tract. The highest value among all tracts for every subject was then selected as the upper limit. These values were utilized in combination with the subject’s brain volume to derive the equations presented below.

\begin{equation}  \label{eq:StreamlineLength}
\text{Length}_{min}= \frac{\sqrt[3]{Volume}}{1.6}, \hspace{5mm}
\text{Length}_{max}= \frac{\sqrt[3]{Volume}}{0.55}
\end{equation}

\subsection{Evaluation}
\label{sec:Evaluation}

We evaluated the accuracy of the automatic tissue segmentation method in terms of Dice similarity coefficient (DSC), 95\textsuperscript{th} percentile of the Hausdorff distance (HD95), and average symmetric surface distance (ASSD). An expert also visually inspected the segmentation maps computed by the automatic method.

To assess the performance of the new anatomically constrained tractography method, we applied it to compute whole-brain tractograms for the 11 independent test subjects. For each brain, we generated a total of 5 million streamlines. Two neuroradiologists with specialized training in pediatric neuroradiology and six years of experience in fetal imaging visually inspected the whole brain tractograms. An expert manually extracted 11 tracts in TrackVis \cite{wang2007diffusion} using the parcellation segmentation labels as ROIs and following the definitions provided by Wassermann et al. \cite{wassermann2016white}. This method required gray matter parcellations. We used an atlas-based procedure to generate parcellation labels using parcellations designed to reflect the ENA33 atlas \cite{blesa2016parcellation}. To achieve this, we utilized ANTs' symmetric deformable registration \cite{avants2008symmetric} to register a previously parcellated T2 atlas of the fetal brain (which can be found at: \url{http://www.crl.med.harvard.edu/research/fetal_brain_atlas/}) to the MD map of the age-corresponding atlas. The extracted tracks were: Anterior Thalamic Radiation (ATR), Genu and Isthmus of the Corpus Callosum (CC 2 and CC 6, respectively), Cortico-Spinal Tract (CST), Frontal Aslant Tract (FAT), Fronto-Pontine Tract (FPT), Inferior Occipito-Frontal fascicle (IFO), Inferior Longitudinal Fascicle (ILF), Middle Longitudinal Fascicle (MLF), Optic Radiations (OR), and Uncinate Fascicle (UF).

To compare our findings with previously established techniques, we employed the Fiber Assignment by Continuous Tracking (FACT) algorithm \cite{mori1999three} to compute whole brain tractography for the same 11 test subjects using the computed diffusion tensor map. We set a whole brain mask based on the segmentation generated by the deep neural network and a step size of 0.6mm for streamline propagation. We manually extracted the tracts from the FACT-computed whole-brain tractograms by following the approach described above and maintaining consistency with the ROIs. 

Subsequently, two neuroradiologists independently assessed each tract generated by our proposed method and FACT using TrackVis. The analysis was conducted in a head-to-head comparative manner, and the radiologists were blinded to which method corresponded to each visualization. However, the visual appearance of the streamlines may have enabled them to distinguish between the methods. The readers were provided with mean diffusivity and fractional anisotropy maps, and no limitations were imposed on window-level setting adjustments. We asked the raters to indicate which visualization was better, significantly better, or if they were of equal quality. We then converted their ratings to a 5-point scale, where 0 indicated no difference between the methods, $\pm1$ indicated that one method was slightly better than the other, and $\pm2$ indicated that one method was significantly better. In our conversion, positive values favor our method, and negative values favor FACT.

We also compared our proposed method with FACT in terms of sensitivity to the angle threshold, which is one of the main settings in every tractography algorithm. Tractography results, especially for the fetal brain, can be highly sensitive to angle threshold setting, requiring careful adjustment to ensure reconstruction of the tracts of interest. Therefore, a method that is less sensitive to this setting is highly desirable as it can improve the tractography accuracy and reduce the need for manual adjustment of the angle setting for every subject/tract. In order to perform this comparison, we applied our method with three angle thresholds of $15^{\circ}$, $20^{\circ}$, and $25^{\circ}$, and we applied FACT with three angle thresholds of $25^{\circ}$, $30^{\circ}$, and $35^{\circ}$. These angles were selected to ensure the best overall tractography accuracy for the two methods. We converted the tracts to binary masks by following these steps: (1) Compute tract density maps, where each voxel shows the number of streamlines crossing that voxel; (2) Compute a density threshold as the 1\% percentile of the non-zero densities; (3) Exclude voxels where the tract density is below the threshold. Other non-zero voxels represent the binary mask of the extracted tract. After computing the binary masks, we computed pair-wise DSC, HD95, ASSD, and relative volume difference (VolDiff) between the three tracts reconstructed with the three angle thresholds to quantify the sensitivity of the method to angle threshold. We define VolDiff between two binary masks $M_1$ and $M_2$ as $\big| \text{vol}(M_1) - \text{vol}(M_1) \big| / \big( \big| \text{vol}(M_1) + \text{vol}(M_1) \big| /2 \big)$.

\section{Results}

\subsection{Assessment of automatic brain tissue segmentation}

\begin{figure}[!ht]
\centering
\includegraphics[width=150mm]{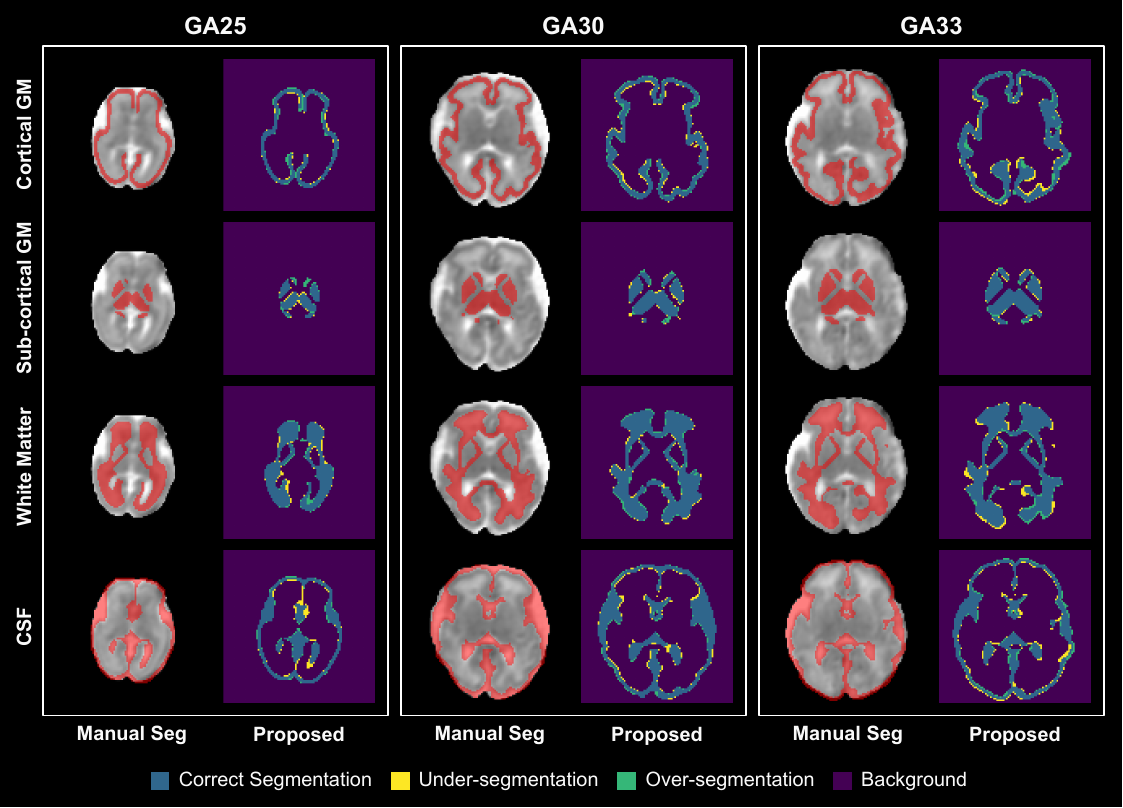}
\caption{Examples of segmentations predicted by our proposed method in three subjects at three different gestational ages (GA). Each panel shows the manual segmentation and the segmentation computed by the proposed method for four tissue types used in the five-tissue-type image. Please note that the last tissue type, reserved for pathological tissue \cite{smith2012anatomically}, is omitted from this figure and excluded from this study, as it is absent in our cohort of healthy fetuses. GM stands for gray matter, and CSF stands for corticospinal fluid.}
\label{fig:tissue_seg_results}
\end{figure}

Figure \ref{fig:tissue_seg_results} shows example fetal brain tissue segmentation results computed by the deep learning method compared with the manually-generated ground truth. They show that the automatic method can accurately segment all four tissue types. Visual assessments by an expert confirmed that the computed segmentations were free of any major errors. Table \ref{table:segmentation_performance_metrics} presents quantitative segmentation performance metrics for the four tissue types. The average DSC for white matter, cortical gray matter, sub-cortical gray matter, and CSF was, respectively, 0.898, 0.830, 0.837, and 0.841. These values are in the range of performance metrics reported for the state-of-the-art segmentation methods applied on fetal T2 MRI in recently-published studies \cite{payette2023fetal, fidon2021distributionally, dou2020deep, dumast2021segmentation}. For white matter, for example, the DSC values achieved by several deep learning methods in a recent study \cite{payette2023fetal} were in the range 0.85-0.90, while for our method in this work it was $0.898 \pm 0.024$. Similarly, surface distance error quantified in terms of HD95 and ASSD are comparable with the results of recent deep learning methods for T2 segmentation \cite{payette2023fetal}, including our own work \cite{karimi2023learning}. These results indicate that the proposed method is able to segment the fetal brain tissue with satisfactory accuracy in dMRI, which has an inherently lower signal quality than T2.

\begin{table*}[!htb]
\centering
\caption{\small{Segmentation performance metrics for the proposed method. Note that higher values of DSC and lower values of HD95 and ASSD are desirable.}}
\label{table:segmentation_performance_metrics}
\begin{tabular}{ L{20mm} C{25mm} C{25mm} C{25mm} C{30mm}  }
\thickhline
& cortical gray matter & sub-cortical gray matter & white matter & cerebrospinal fluid \\ \thickhline
DSC       & $0.830 \pm 0.034$ & $0.837 \pm 0.045$ & $0.898 \pm 0.024$ & $0.841 \pm 0.030$ \\
HD95 (mm) & $1.049 \pm 0.022$ & $1.051 \pm 0.030$ & $0.955 \pm 0.028$ & $1.047 \pm 0.025$ \\
ASSD (mm) & $0.240 \pm 0.110$ & $0.235 \pm 0.120$ & $0.209 \pm 0.126$ & $0.246 \pm 0.133$ \\
\thickhline
\end{tabular}
\end{table*}

\subsection{Assessment of extracted tracts}

\begin{figure}[!ht]
\centering
\includegraphics[width=150mm]{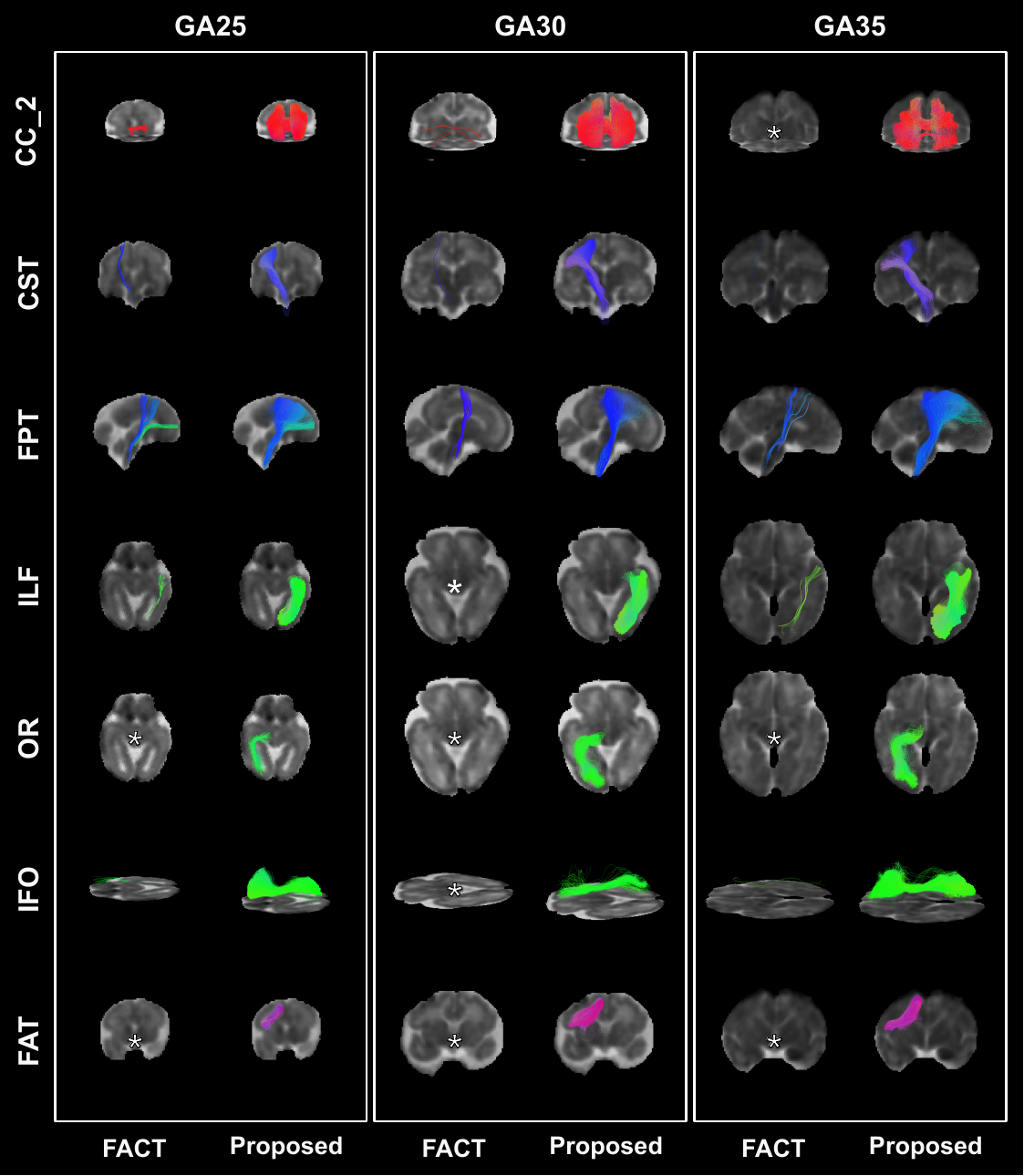}
\caption{A visual comparison of seven different tracts reconstructed using the fiber assignment by continuous tracking (FACT) algorithm and our proposed method for three fetal brains at different gestational ages (GA). The following tracts are shown: Genu of the Corpus Callosum (CC\_2), Cortico-Spinal Tract (CST), Frontal Aslant Tract (FAT), Fronto-Pontine Tract (FPT), Inferior Occipito-Frontal fascicle (IFO), Inferior Longitudinal Fascicle (ILF), and Optic Radiations (OR). A white asterisk on top of the image means the tract was not reconstructed by that method.}
\label{fig:tractography_results}
\end{figure}

All 11 tracts were successfully extracted from the whole-brain tractograms generated with our proposed method and 10 with FACT. It was observed that the FAT tract was consistently present in the tractograms generated with our proposed method but was absent in the tractograms generated with FACT. Therefore, this tract was excluded for further analysis. Figure \ref{fig:tractography_results} displays seven selected tracts for a visual comparison of the two methods. As shown in this figure, FACT failed to reconstruct all tracts consistently, while our method successfully reconstructed all of them. Figure \ref{fig:tractography_results} summarizes the results of a head-to-head comparison of our proposed method and FACT by two experts. It shows that the new anatomically constrained tractography method was consistently superior to FACT in terms of the tract reconstruction success rate and the scores provided by the experts.

Table \ref{table:tract_recon_angle} summarizes the results of the experiment to quantify the impact of angle threshold setting on the reconstruction of the tracts for the proposed method and FACT. In this experiment, the tracts were converted to binary masks, as described in Section \ref{sec:Evaluation} and several metrics of agreement/disagreement between pairs of tracts reconstructed with different angle thresholds were computed. The values in Table \ref{table:tract_recon_angle} represent the summary for all 11 tracts considered. These results show that the new method is highly robust to the setting. Changing the angle threshold between $15^{\circ}$ and $25^{\circ}$ had a small impact on the tract reconstruction results, with a mean pair-wise DSC of 0.941 for all tracts. For all four evaluation criteria shown in Table \ref{table:tract_recon_angle}, the results for our method were significantly ($p<0.001$) better compared with FACT.

\begin{table*}[!htb]
\centering
\caption{Results of the experiment to assess the robustness of tractography methods to angle threshold setting. The values shown are computed pair-wise between the tracts reconstructed with different angle thresholds. Note that higher values of DSC and VolDiff and lower values of HD95 and ASSD are desirable.}
\label{table:tract_recon_angle}
\begin{tabular}{ L{30mm} C{25mm} C{25mm} C{25mm} C{25mm} }
\thickhline
& DSC & HD95 (mm) & ASSD (mm) & VolDiff \\ \thickhline
Proposed method   & $0.941 \pm 0.042$ & $0.815 \pm 0.027$ & $0.208 \pm 0.103$ & $0.083 \pm 0.032$ \\
FACT              & $0.605 \pm 0.119$ & $1.664 \pm 0.157$ & $0.751 \pm 0.195$ & $0.279 \pm 0.122$ \\
\thickhline
\end{tabular}
\end{table*}

\begin{figure}[!ht]
\centering
\includegraphics[width=100mm]{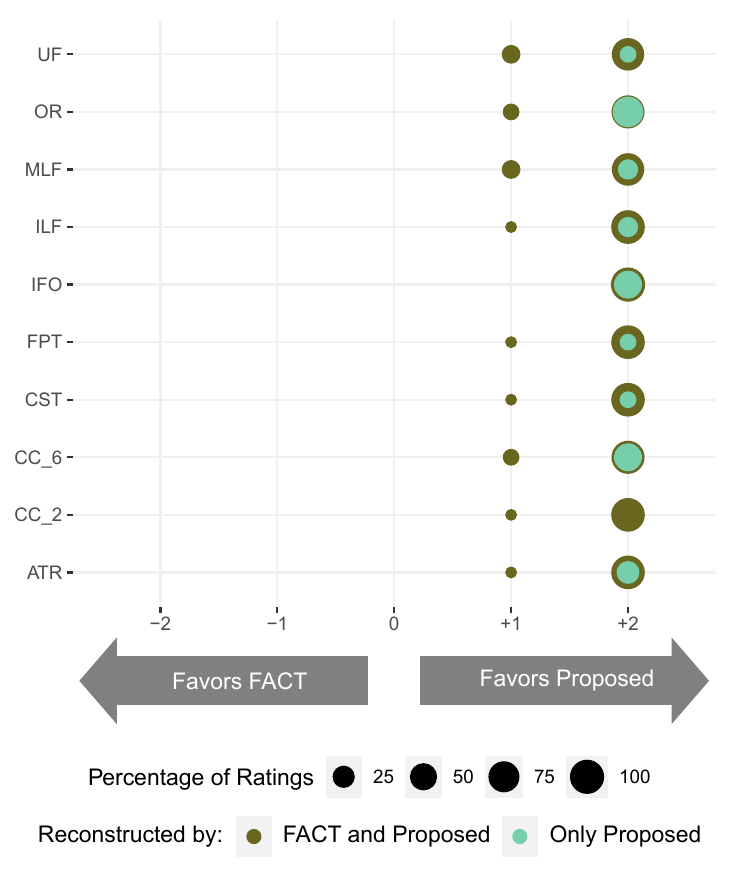}
\caption{Head-to-head comparison of different tracts reconstructed with our proposed method and FACT, depicted using a balloon plot. The tracts were extracted by following the same procedure for both our anatomically constrained tractography method and FACT. Each circle's size indicates the percentage of cases assigned a given score. These results were computed from 11 fetuses (one per gestational week between 23 and 35 weeks) and two raters. Negative scores (left side of the plot) favor FACT, and positive scores (right side) favor the proposed method. Brown balloons indicate tracts scored when both FACT and our proposed method could reconstruct the tracts, while green balloons represent tracts reconstructed exclusively by our proposed method. The following tracts were used for the evaluation: Anterior Thalamic Radiation (ATR), Genu (CC\_2) and Isthmus (CC\_6) of the Corpus Callosum, Cortico-Spinal Tract (CST), Fronto-Pontine Tract (FPT), Inferior Occipito-Frontal fascicle (IFO), Inferior Longitudinal Fascicle (ILF), Middle Longitudinal Fascicle (MLF), Optic Radiation (OR), and Uncinate Fascicle (UF).}
\label{fig:seg_results}
\end{figure}

\section{Discussion}

This paper introduces the development and validation of an anatomically constrained fetal brain tractography method. Our method relies on the classification/segmentation of brain tissue into white matter, cortical and subcortical gray matter, and cerebrospinal fluid. These segmentations are automatically computed by a deep learning method directly in the dMRI space, a valuable capability that has not been reported in prior works. The segmentation method only requires the diffusion tensor map as the input. The same diffusion tensor map is used to infer the streamline propagation direction for computing the tractogram. We convert the diffusion tensor into a sharpened diffusion orientation distribution function that helps minimize the probability of computing spurious streamlines and enables advanced probabilistic tractography methods such as iFOD2.

Our automatic segmentation method outperforms multi-atlas-based approaches. In our previous work, we achieved a DSC of $0.75 \pm 0.17$ for cortical gray matter, $0.69 \text{--} 0.86 \pm 0.14$ for sub-cortical gray matter structures, and $0.70 \text{--} 0.87 \pm 0.15$ for white matter structures \cite{calixto2023detailed}. The new deep learning method far surpasses this method, with higher average DSC values and narrower standard deviations for all tissue types. Moreover, multi-atlas-based automatic segmentation for this data typically takes 15-25 minutes per subject, while our deep learning approach usually has an inference time of 3-5 seconds. Surface distance errors presented in Table \ref{table:segmentation_performance_metrics} show that HD95 (representing large surface distance errors) for white matter, cortical gray matter, and sub-cortical gray matter are all approximately 1mm, while the ASSD (representing average surface distance errors) are approximately 0.20-0.25 mm. These numbers suggest that our method can be used for accurate seeding and stopping operations that are essential for anatomically constrained tractography.

Accurate tissue classification enables us to: (1) Launch the streamlines at seed points on the boundary between white matter and gray matter, and (2) Effectively remove the spurious and anatomically invalid streamlines and only preserve valid streamlines based on their paths and endpoints. Effective use of anatomical information enables us to avoid premature termination of valid streamlines or generate implausible streamlines that are prevalent in fetal tractography, as observed in this study with FACT. Standard tractography methods rely purely on the local diffusion tensor information for streamline propagation and termination (e.g., based on FA). This can lead to high false positive and false negative rates because of the low fetal dMRI data quality that can give rise to noisy and inaccurate diffusion tensor estimates.

We evaluated the new method on a set of independent test subjects and compared the results to tractography generated using FACT. Our manual extraction of the tracts followed established criteria outlined by Wassermann et al. \cite{wassermann2016white} regarding where streamlines should start and pass through and which structures to avoid. We followed the same tract definitions and extraction procedures for both tractography methods. Expert evaluations showed that our method consistently outperformed FACT in terms of expert scores. Our method was consistently successful in reconstructing prominent white matter tracts. A visual comparison of the tracts reconstructed by our method and the results produced from the Human Connectome Project data \cite{wasserthal2018tractseg} show that our results on fetal brains are qualitatively comparable with adult brains.

While FACT was able to reconstruct most tracts (except for the FAT) as evidenced in Figure \ref{fig:seg_results}, the tractography results generated by FACT were often limited in their ability to reconstruct a sufficient number of streamlines. While some cases in our method (like the FPT track of the GA30 fetus in Figure \ref{fig:tractography_results}) exhibited less dense frontal fibers than anticipated, our approach utilizes a probabilistic algorithm that can be remedied by increasing the number of streamlines. Conversely, increasing the number of streamlines in FACT would have no impact, as the pathways of the streamlines are already predetermined by the principal eigenvector of the tensor and, thus, would not improve the reconstruction of empty areas. Furthermore, our approach successfully reconstructed tracts with highly curved areas like the optic radiations and obtained the lateral projections of the CST.

A comparison of our results with the results of prior works on in-utero fetal tractography exhibits significant advancements in the anatomical accuracy and completeness of neural tracts, addressing limitations observed in earlier research. Here, we detail these comparisons for specific tracts, underscoring the enhancements our method provides. It's worth noting that comparisons for the ATR, FAT, and MLF are absent, as no previous group has extracted them on fetuses.

\begin{itemize}

\item \textbf{Corpus Callosum:} Prior investigations have focused on various segments of the corpus callosum \cite{jaimes2020vivo,machado2021spatiotemporal,wilson2021development,kasprian2008utero,zanin2011white,mitter2015vivo}. Our approach stands out in its ability to consistently generate tracts connecting both the medial and lateral surfaces of the frontal lobes within the forceps minor. Furthermore, our streamlines terminate accurately at the gray-white matter interface without prematurely ending or crossing the cortical plate and ending into the CSF.

\item \textbf{CST:} Earlier studies have encountered difficulties in fully extracting the CST, often resulting in premature termination of streamlines \cite{kasprian2008utero,zanin2011white} or failure to capture lateral projections \cite{jaimes2020vivo,machado2021spatiotemporal}. Unlike these studies, our approach consistently captures the complete intracranial CST, avoiding inaccuracies such as streamline termination in the CSF or inappropriate crossing through deep gray matter structures \cite{wilson2021development}.

\item \textbf{FPT:} Limited research has depicted the FPT in fetuses. Our findings show more extensive tract delineation, reaching all intended frontal targets without the premature termination observed in previous work \cite{mitter2015vivo}.

\item \textbf{IFO:} Unlike other works, our tracts reach all expected cortical targets \cite{song2015asymmetry,jaimes2020vivo,machado2021spatiotemporal}, terminate at the junction of gray and white matter\cite{mitter2015vivo} and most importantly, traverse the expected path through the external capsule.

\item \textbf{ILF:} Our work surpasses previous works \cite{song2015asymmetry, jaimes2020vivo} on generating a fuller tract that reaches all cortical targets and has results comparable to those of Machado et al. \cite{machado2021spatiotemporal} and Wilson et al. \cite{wilson2021development} However, it is worth noting that in the two works above, only some branches of the ILF are visualized—e.g., missing one connecting the cuneus to the anterior mesial temporal gyri.

\item \textbf{OR:} Unlike previous attempts, our method ensures that optic radiation streamlines do not prematurely terminate \cite{zanin2011white} and exclusively traverse white matter to reach the occipital lobe, avoiding aberrant terminations \cite{wilson2021development} or ventricular crossings \cite{corroenne2023tractography} noted in earlier studies.

\item \textbf{UF:} While our findings align with the trajectory observed in previous studies for the UF \cite{jaimes2020vivo,mitter2015vivo}, we additionally note that our tracts achieve expected orbital targets, indicating a refinement over prior work.

\end{itemize}

Our study has some limitations. Firstly, the sample size of 84 fetuses is relatively small due to the challenges of subject recruitment, fetal imaging, and the moderate success rate of the dMRI reconstruction pipeline. Moreover, generating manual segmentation labels on 3D images is very time-consuming. We evaluated our tissue segmentation method and the complete tractography technique on 11 independent test subjects and found statistically significant improvements compared with FACT. Nonetheless, increasing the data size for method development and validation will be useful. Secondly, our data was obtained from a single center, and we employed an in-house pipeline for preprocessing of the dMRI data. Validation of data from other centers helps assess the generalizability of our method. Additionally, evaluating our method with higher quality or multi-shell dMRI data can also be interesting. As we have pointed out above, unlike some recent works that have used unique, high-quality data from the Human Connectome Project \cite{wilson2023spatiotemporal, wilson2021development}, our method has been designed to work with more typical dMRI scans. We think this is a significant advantage of our method because HCP-style data are very rare in fetal studies. Nonetheless, applying our method to higher-quality data will be useful in highlighting the potential advantages and limitations of our method compared with existing techniques. Another limitation of our evaluations was that our data included healthy subjects only. Future works can study the performance of our method on abnormal fetal brains. Lastly, we anticipate that our method can lead to significant improvements in the accuracy and reproducibility of structural connectivity analysis, which is a main application of tractography. We aim to pursue this line of investigation in our future work.

\section{Conclusions}

Our study has demonstrated the feasibility of anatomically-constrained tractography of the fetal brain. The new method proposed in this work presents a significant stride toward improving the accuracy of fetal brain tractography. The higher tract reconstruction success rate and accuracy offered by our method can enhance the accuracy of dMRI-based studies of the fetal brain white matter in utero. It can also enable reliable quantitative assessment of the structural connectivity of the fetal brain. Ultimately, these new capabilities can enable important scientific and clinical applications and lead to a better understanding of brain development at its earliest stage.

\section*{Acknowledgements}

This research was supported in part by the National Institute of Neurological Disorders and Stroke under award numbers R01NS128281 and R01 NS106030; the Eunice Kennedy Shriver National Institute of Child Health and Human Development under award numbers R01HD110772 and R01 HD109395; the National Institute of Biomedical Imaging and Bioengineering under award numbers R01 EB031849, R01 EB032366, R01 EB018988, and R01 EB013248; the Office of the Director of the NIH under award number S10 OD0250111; the Rosamund Stone Zander Translational Neuroscience Center, Boston Children's Hospital; the Office of Faculty Development at Boston Children's Hospital; and a scholarship from the American Roentgen Ray Society. The content of this publication is solely the responsibility of the authors and does not necessarily represent the official views of the NIH.

\bibliographystyle{unsrt}
\bibliography{davoodreferences}

\end{document}